\documentclass{article}
\usepackage{spconf,amsmath,epsfig}
\usepackage{amssymb}
\usepackage{amsmath}
\usepackage{amsfonts}
\usepackage{booktabs}

\def\x{{\mathbf x}}
\def\y{{\mathbf y}}
\def\z{{\mathbf z}}
\def\u{{\mathbf u}}

\def\p{{\mathbf p}}
\def\L{{\cal L}}

\title{SegSALSA-STR: A convex formulation to supervised hyperspectral image segmentation using hidden fields and structure tensor regularization}
\name{ Filipe  Condessa$~^\textbf{a,b,c,e}$, Jos{\'e}
  Bioucas-Dias$~^\textbf{a,b}$,  and Jelena Kova{\v
    c}evi{\'c}$~^\textbf{c,d,e}$\thanks{
This paper was submitted to IEEE WHISPERS 2015: 7$^\textrm{th}$
Workshop on Hyperspectral Image and Signal Processing: Evolution on
Remote Sensing. The authors gratefully acknowledge support from the Portuguese Science and Technology Foundation under projects UID/EEA/50008/2013, PTDC/EEI-PRO/1470/2012, the Portuguese Science and Technology Foundation and the CMU-Portugal (ICTI) program under grant  SFRH/BD/51632/2011, NSF through award 1017278, and the CMU CIT Infrastructure Award.
}}
\address{$^\textbf{a}$Instituto de Telecomunica\c{c}\~oes, Lisboa, Portugal \\
$^\textbf{b}$Instituto Superior T\'ecnico, University of Lisbon, Lisboa, Portugal\\
$^\textbf{c}$Department of ECE, Carnegie Mellon University, Pittsburgh, PA, USA\\
$^\textbf{d}$Department of BME, Carnegie Mellon University, Pittsburgh, PA, USA\\
$^\textbf{e}$Center for Bioimage Informatics, Carnegie Mellon University, Pittsburgh, PA, USA
}
\begin{document}
%
\maketitle

\begin{abstract}
In this paper we present a supervised hyperspectral image segmentation algorithm based on a convex formulation of a marginal maximum \emph{a posteriori} segmentation with hidden fields and structure tensor regularization: Segmentation via the Constraint Split Augmented Lagrangian Shrinkage by Structure Tensor Regularization (SegSALSA-STR).
This formulation avoids the generally discrete nature of segmentation problems and the inherent NP-hardness of the integer optimization associated.

We extend the Segmentation via the Constraint Split Augmented
Lagrangian Shrinkage (SegSALSA) algorithm~\cite{BioucasDiasCK:14} by generalizing the vectorial total variation prior using a structure tensor prior constructed from a patch-based Jacobian~\cite{lefkimmiatis2013convex}.
The resulting algorithm is convex, time-efficient and highly
parallelizable.
This shows the potential of combining hidden fields with convex
optimization through the inclusion of different regularizers.
The SegSALSA-STR algorithm is validated in the segmentation of real hyperspectral images.
\end{abstract}

\section{Introduction}
\label{sec:intro}
Supervised image segmentation is fundamental in a large number of hyperspectral image applications~\cite{bioucas2013hyperspectral}.
Image segmentation aims to partition the image domain such that pixels belonging to the same partition element share similar properties, either by raw pixel values or more complex features.
Due to its ill-posed nature (\emph{e.g.} depending on the goal, multiple similarity criteria for grouping pixels in the same partition element can be used), image segmentation is often performed with help of regularization that promote or penalize different behaviors, \emph{e.g.} the use of a Markov Random Field~\cite{li1995markov} MRF) to promote smoothness of the labeling.

As partitions are naturally represented by images of integers, maximum
\emph{a posteriori} (MAP) segmentations become integer optimization
problems. SegSALSA~\cite{BioucasDiasCK:14,CondessaBDK:14} sidesteps
the discrete nature of the segmentation problems by using a hidden
field to drive the segmentation and marginalizing on the hidden
field~\cite{marroquin2003hidden} combined with a vectorial total
variation (VTV) prior~\cite{goldluecke2012natural,sun2013classTV}. The
segmentation is inferred by computing the marginal maximum \emph{a
  posteriori} (MMAP) estimate of the hidden field, which is a convex
problem solved by the Split Augmented Lagrangian Shrinkage (SALSA) algorithm~\cite{afonso2011augmented}.

In this paper, we extend the formulation of SegSALSA to include a generalization of VTV prior based on structure tensor priors~\cite{lefkimmiatis2013convex}, thus deriving a more general algorithm for hyperspectral image segmentation.
The structure tensor prior is constructed from a patch-based Jacobian.
The regularization via the structure tensor prior is achieved via a
Schatten norm regularization, akin to~\cite{lefkimmiatis2012hessian}.

The paper is organized as follows.
Section \ref{sec:background} describes the SegSALSA algorithm, introducing the hidden fields and the marginal MAP formulation.
Section \ref{sec:prior} presents the structure tensor regularization and describes the construction of the structure tensor prior from the patch-based Jacobian.
Section \ref{sec:optimization} formulates our optimization problem and presents the SegSALSA-STR algorithm.
Section \ref{sec:results} illustrates the validity of the algorithm on synthetic and real hyperspectral data, and Section \ref{sec:conclusion} concludes the paper.
\section{Background}
\label{sec:background}
 Let $x\in\mathbb{R}^{d\times n}$ denote a $n$-pixel hyperspectral image with $d$ spectral bands, where $\x_i \in\mathbb{R}^{d}$ denotes the feature vector corresponding to the image pixel $i$, $\mathcal{S} = \{1, \hdots, n\}$ the set that indexes the image pixels, $\mathcal{L} = \{1, \hdots, K\}$ the set of possible $K$ labels, and $\y \in \mathcal{L}^n$ a labelling of the image.

\paragraph*{Maximum \emph{a posteriori} segmentation}
The SegSALSA formulation adopts a Bayesian perspective; the MAP segmentation associated with the labeling $\widehat{\y}$ is obtained as
\begin{equation}
\label{eq:MAP}
\widehat{\y} = \arg\max_{\y \in \cal L ^N} p(\y|\x) = \arg\max_{\y \in
  \L ^N} p(\x| \y) p(\y),
\end{equation}
where $p(\y|\x)$ and $p(\x|\y)$ denote the posterior probability and the observation model respectively, and $p(\y)$ the prior on the labeling.
Under assumption of conditional independence of the observation model, we can expand the observation model
$p(\x|\y) = \prod_{i \in \cal S} p(\x_i | y_i).$
The optimization associated MAP formulation \eqref{eq:MAP} is an integer optimization problem:
apart from a small number of exceptions, the use of contextual priors $p(\y)$, such as MRF priors makes \eqref{eq:MAP} a combinatorial problem.

\paragraph*{Hidden fields and marginal MAP}
To mitigate the difficulties associated with the integer optimization problem of the MAP, we move from the discrete formulation by introducing a hidden field~\cite{marroquin2003hidden} and marginalizing on the discrete labels.
We represent the hidden field $\z$ as a $K\times n$ matrix containing a collection of hidden vectors $\z_i \in \mathbb{R}^K$ for $i \in \cal{S}$.
The joint probability of the labeling $\y$ and the field $\z$ is $p(\y,\z) = p(\y|\z)p(\z)$, with assumption of conditional independence of $p(\y|\z) = \prod_{i \in \cal S} p(y_i|\z_i)$.
The joint probability of the features $\x$, the labeling $\y$ and the field $\z$ is $p(\x,\y,\z) = p(\x|\y) p(\y | \z) p(\z)$.
We can now marginalize on the discrete labels,
\begin{equation*}
p(\x,\z) = \prod_{i\in\cal S} \big\{ \sum_{y_i \in \cal L} p(\x_i,y_i) p(y_i, \z_i) \big\} p(\z),
\end{equation*}
and obtain a marginal MAP (MMAP) estimate of the hidden field,
\begin{align}
\label{eq:pre_mmap}
\widehat{\z}_\textrm{MMAP} & = \arg \max_{\z \in \mathbb{R}^{K \times n}} p(\x,\z) \nonumber \\
&=  \arg \min_{\z \in \mathbb{R}^{K \times n}} -\ln p(\x|\z) - \ln p(\z),
\end{align}
which is no longer a discrete optimization problem.

\paragraph*{Link between class labels and hidden fields}
As a  model for the conditional probabilities $p(y_i| \z_i )$ for the labels given the field, we adopt
\begin{equation*}
p(y_i = k | \z_i) \equiv [\z_i]_k,
\end{equation*}
for $i \in \cal S$ and $k \in \cal L$.
This link between the probabilities of the class labels and the hidden fields imposes two constraints on the hidden field and, consequently, on the optimization \eqref{eq:pre_mmap}: the field must be nonnegative, and each hidden vector $\z_i$ for $ i \in \cal S$ must sum to one.
We can now formulate the optimization \eqref{eq:pre_mmap} as
\begin{align}
\label{eq:mmap}
\widehat{\z}_\textrm{MMAP} =  \arg \min_{\z \in \mathbb{R}^{K \times n}} -\ln p(\x|\z) - \ln p(\z), \\
\textrm{subject to:  } \quad \quad \z \geq 0, \quad \quad \textbf{1}_K^T \z = \textbf{1}_n .\nonumber
\end{align}
\section{Structure Tensor Regularization}
\label{sec:prior}
We extend the SegSALSA algorithm by replacing the VTV prior by a generalization of VTV based on  structure tensor regularization~\cite{lefkimmiatis2013convex}.

\paragraph*{Patch-based Jacobian}
Following closely the notation in~\cite{lefkimmiatis2013convex}, we
define the patch-based Jacobian of the hidden field as 
\begin{equation}
\label{eq:patch}
[J \z]_i^T = \begin{bmatrix}  (P_1 D_h \z)_i ^T &\hdots & (P_L D_h
\z) _i^T  \\
 (P_1 D_v \z) _i^T &\hdots & (P_L D_v \z ) _i^T  \end{bmatrix},
\end{equation}
where $[J \z]_i$ denotes the components of the
patch-based Jacobian on the $i$th pixel of the hyperspectral image,
corresponding to a $( K L) \times 2$ matrix. 
 $D_h$ and $D_v$ are the horizontal and vertical difference operators,
 and $P_j$ is a weighted shift operator corresponding to the patch,
with each operator being applied equally to the entire field $D_h, D_v, P_j :\mathbb{R}^{K \times n} \rightarrow \mathbb{R}^{K \times n} $.
Assuming the patch to be a $(2 M + 1) \times (2 M + 1)$ rectangular patch, with $L =  (2 M + 1)^2$ pixels, $P_j$ corresponds to the $j$th possible shift within the patch (from the $L$ possible shifts) weighted by a Gaussian centered on the center of the patch and with a bandwidth $\gamma$.

\paragraph*{Discrete structure tensor}
From the patch-based Jacobian of the hidden field \eqref{eq:patch},
the structure tensor $S_L$ is defined as
\begin{equation}
\label{eq:stensor}
[S_L \z]_i = [J \z]_i^T [J \z]_i,
\end{equation}
a $2 \times 2$ matrix for the $i$th pixel of the hyperspectral image.

The minimization of the eigenvalues of the structure tensor $S_L$ in \eqref{eq:stensor}  leads to the penalization of variations of the field among the pixels in patch.
As there is an intrinsic connection between the eigenvalues of the
structure tensor 
\eqref{eq:stensor} $(\lambda_+, \lambda_-)$
and the singular values of the patch-based Jacobian
$(\sqrt{\lambda_+},\sqrt{\lambda_-})$, we can minimize the singular
values instead.
Let $\|[J \z]_i \|_{S_p}$ denote the Schatten $p$ norm of the patch-based Jacobian
\begin{equation*}
\|[J \z]_i \|_{S_p} = \| \sigma([J \z]_i) \|_p,
\end{equation*}
where $\sigma([J \z]_i)$ represent the singular values of $[J \z]_i$.
The discrete structure tensor prior can be constructed through the
minimization of the singular values of the patch-based Jacobian
\begin{equation}
\label{eq:prior}
- \ln p(\z) \equiv   \lambda \sum_{i \in \mathcal{S}} \|[J \z]_i \|_{S_p}  + c^{te}.
\end{equation}
It should be noted that for $1 \times 1$ patches  ($L = 1$), the minimization of the Schatten norm of the structure tensor is equivalent to the minimization of the VTV, leading to the SegSALSA formulation in~\cite{BioucasDiasCK:14,CondessaBDK:14}.
\section{Optimization Algorithm}\label{sec:optimization}
With the SegSALSA general formulation described in Section
\ref{sec:background} and armed with the new generalized total variation prior from
the structure tensor described in Section \ref{sec:prior} we can now
describe our algorithm which we name SegSALSA-STR.
\paragraph*{Problem formulation}
Combining the MMAP formulation in \eqref{eq:mmap} with the prior described in \eqref{eq:prior}, we can formulate our MMAP problem as follows:
\begin{align}
\label{eq:problem}
\widehat{\z}_\textrm{MMAP} = & \arg \hspace{-0.1in}\min_{\z \in \mathbb{R}^{K\times n}} \sum_{i \in \cal S} -\ln(\p_i^T \z_i) + \lambda \sum_{i\in \cal S} \| [J \z]_i \|_{S_p}\\
&\textrm{subject to: }\quad\quad \z \geq 0 \quad\quad  \textbf{1}_K^T \z = \textbf{1}_n^T \nonumber,
\end{align}
where $\p_i = [p(\x_i | y_i = 1) \hdots p(\x | y_i =  K)]^T$, and $\p_i^T \z_i > 0 $ in the feasible set.
As the Hessian of $-\ln(\p_i^T \z_i)$ is semidefinite positive, and $\| [J \z]_i \|_{S_p}$ is a composition of norms, the optimization \eqref{eq:problem} is convex.
The solution $\widehat{\z}_\textrm{MMAP}$ is computed by the SALSA methodology~\cite{afonso2011augmented}.
However, we cannot apply the SALSA methodology directly to \eqref{eq:problem}.
To do so we rewrite the problem as a sum of convex functions with linear constraints.

\paragraph*{Rewriting the optimization problem}
We rewrite the optimization problem \eqref{eq:problem} as
\begin{equation}
\label{eq:opt_1}
\min_{\z \in \mathbb{R}^{K\times n}} \sum_{j=1}^4 g_j(H_j \z),
\end{equation}
where $H_j$, for $j = 1, \hdots, 4$, denote linear operators, and $g_k$, for $j = 1, \hdots, 4$ denote closed, proper, and convex functions.
We define the linear operators $H_j$ as:
\begin{equation}
H_1 = I, \quad H_2 = J, \quad H_3 = I, \quad H_4 = I,
\end{equation}
where $I:\mathbb{R}^{K\times n} \rightarrow  \mathbb{R}^{K\times n} $
denotes the identity operator and $J:\mathbb{R}^{K\times n}
\rightarrow  \mathbb{R}^{2 L K\times n}  $ denotes an operator stacking the patch-based Jacobians defined in \eqref{eq:patch}.
We define the closed, proper, and convex functions $g_j$ as:
\begin{align}
g_1(\xi) = \sum_{i \in \cal S} - \ln(\p_i^T \xi),&\quad
g_2(\xi) = \lambda \sum_{i \in \cal S} \| \xi \|_{S_p},  \\
g_3(\xi) = i_+(\xi)  , & \quad g_4(\xi) = i_1(\textbf{1}_K^T\xi). \nonumber 
\end{align}
The functions $i_+$ and $i_1$ are indicator functions for the sets $\mathbb{R}_+^{K\times n}$ and ${\textbf{1}_n}$, respectively.
The introduction of a variable splitting 
\begin{equation*}
\u_j = H_j \z,
\end{equation*}
for $j = 1, \hdots, 4$, and with $ \u \in \mathbb{R}^{(3+  2 L) K\times n}$, allows us to reformulate the optimization \eqref{eq:opt_1} into a constrained formulation
\begin{equation}
\label{eq:constrained}
\min_{\textbf{u},\z} \sum_{j=1}^4 g_j(\u_j), \quad \textrm{subject to } \u = G \z,
\end{equation}
with $G$ the corresponding to the stacking of the linear operators $H_j$, for $j = 1, \hdots, 4$.

\paragraph*{Salsa methodology}
The optimization \eqref{eq:constrained} is solved following the SALSA methodology~\cite{afonso2011augmented}, an instance of the alternating direction method of multipliers designed to optimize sum of an arbitrary number of convex terms.
Let $\mathbf{d}$ denote the scaled Lagrange multipliers; solving \eqref{eq:constrained} reduces to the following iterative decoupled problem
\begin{align}
\label{eq:z_update}
\z^{k+1} = &~ \arg \min_{\z} \|G \z - \u^k -\mathbf{d}^k \|_F^2, \\
\label{eq:u_update}
\u^{k+1} = &~ \arg \min_{\u} \sum_{j=1}^4 g_j(\u_j) + \frac{\mu}{2} \|G \z^k - \u - \mathbf{d}^k \|_F^2, \\
\label{eq:d_update}
\mathbf{d}^{k+1} = &~ \mathbf{d}^k - (G \z^{k+1} - \u^{k+1} ).
\end{align}

\paragraph*{Solving the iterative decoupled problem}
The quadratic problem \eqref{eq:z_update} is solved by computing independent cyclic convolutions on each image of $\z$, with time complexity $O( L K n \log n)$.
The problem \eqref{eq:u_update} can be decoupled for each of the convex functions $g_j$, amounting to computing the Moreau proximity operator (MPO)~\cite{combettes2011proximal} for each of the functions.

For $j = 1,3,4,$ involved in the respective MPO is respective the
MPOs, all with $O(Kn)$ complexity are as follows: $g_1$ is the
root of a polynomial using a closed form solution; $g_3$ is the
projection onto the positive orthant; and $g_4$ is the projection onto
the probability simplex.
For $j = 2$, the MPO is the solution of the following problem
\begin{equation*}
\u^{k+1}_{2} = \arg \hspace{-0.15in}\min_{\hspace{-0.1in}\u_2 \in \mathbb{R}^{(2  L  K)\times n}} \sum_{i \in \cal S} \| [\u_2]_i \|_{S_p} + \frac{\mu}{2 \lambda} \|H_2 \z^k - \u_2 - d^k_2 \|_F^2.
\end{equation*}
For $p=1$, which we adopt in this paper, this can be solved by the soft thresholding of
the singular values of  $H_2 \z^k - d_2^k$.

\begin{figure*}[]
\begin{center}
\begin{tabular}{ccccc}
\includegraphics[height=.35\columnwidth,width=.37\columnwidth]{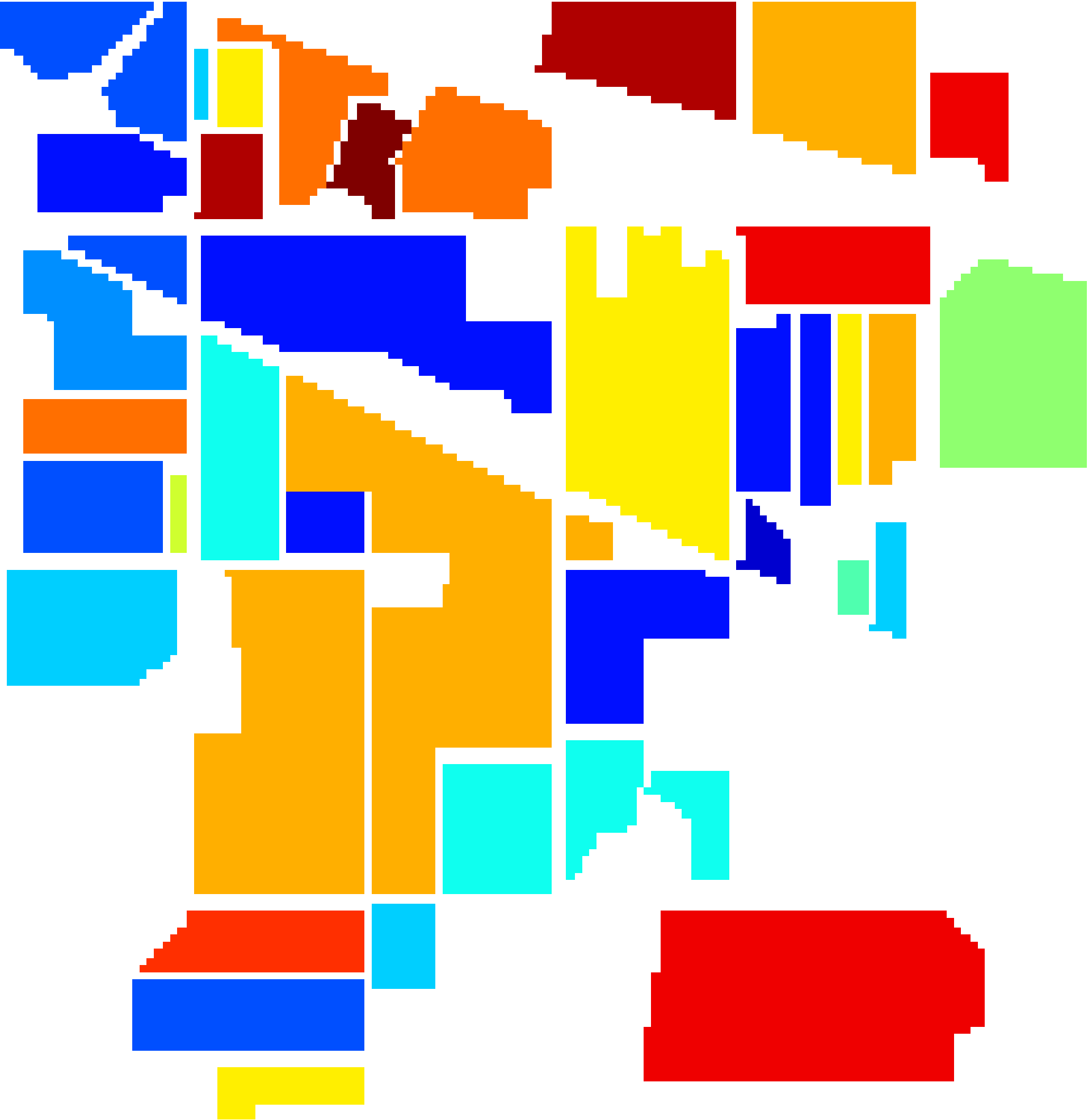} &
\includegraphics[height=.35\columnwidth,width=.37\columnwidth]{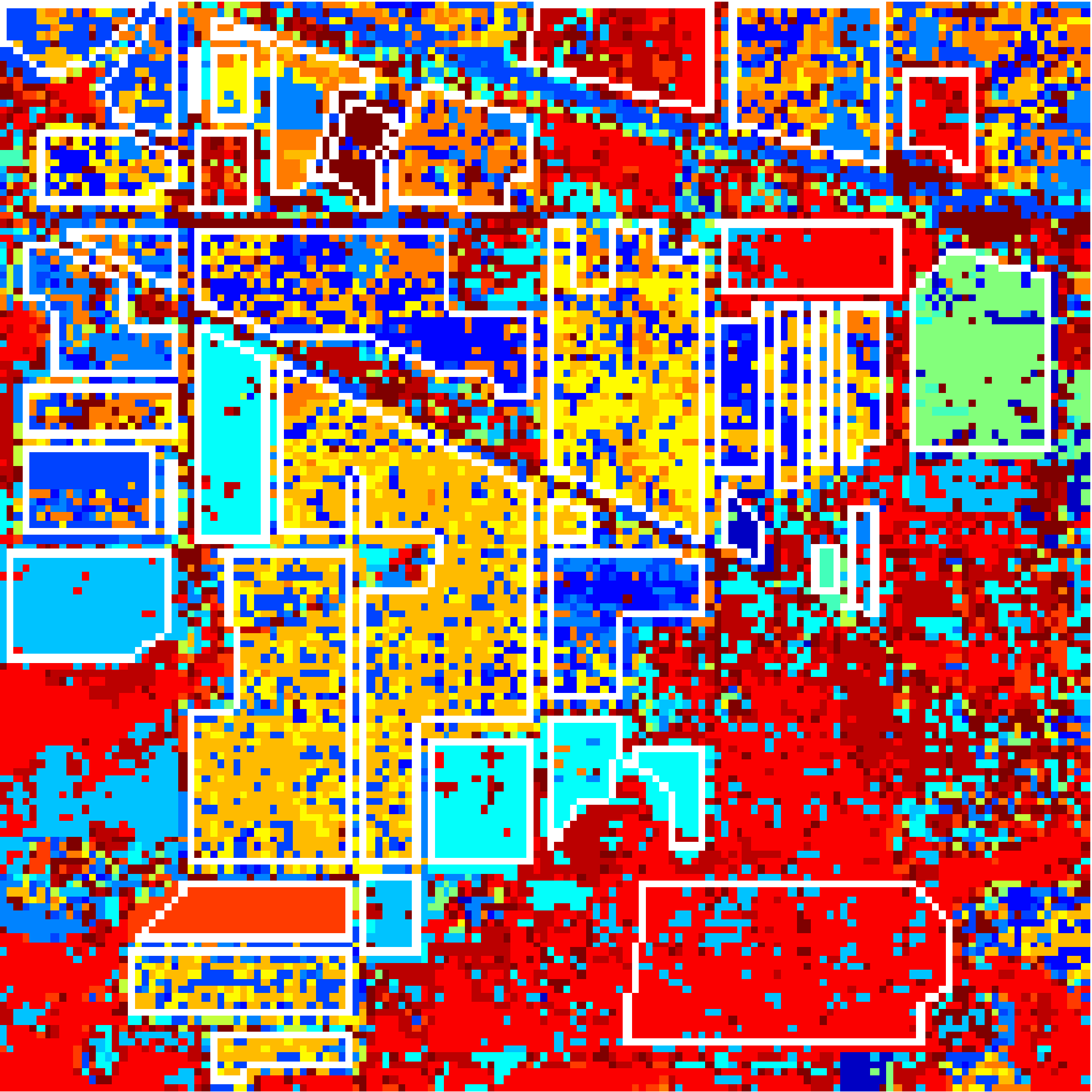} &
\includegraphics[height=.35\columnwidth,width=.37\columnwidth]{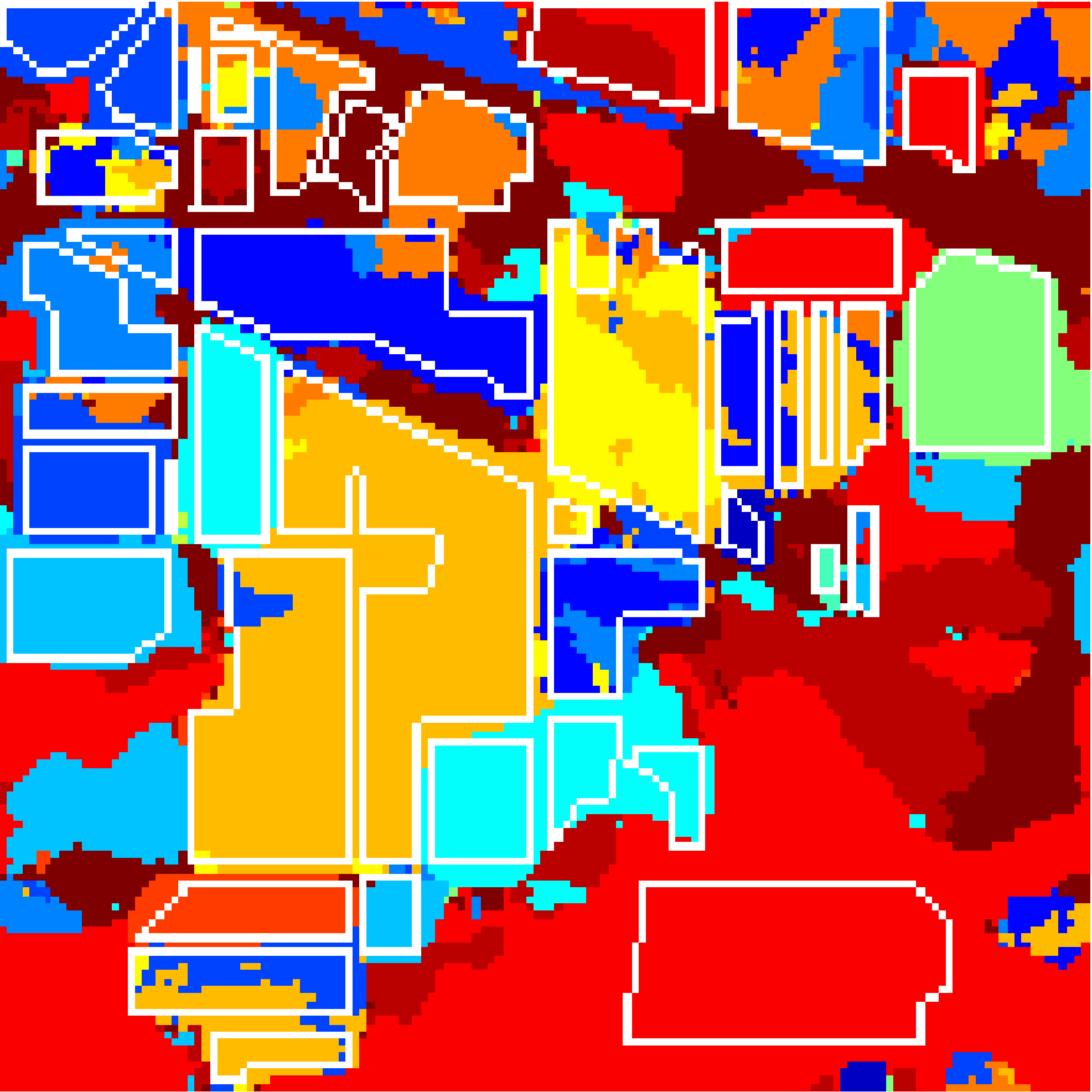} &
\includegraphics[height=.35\columnwidth,width=.37\columnwidth]{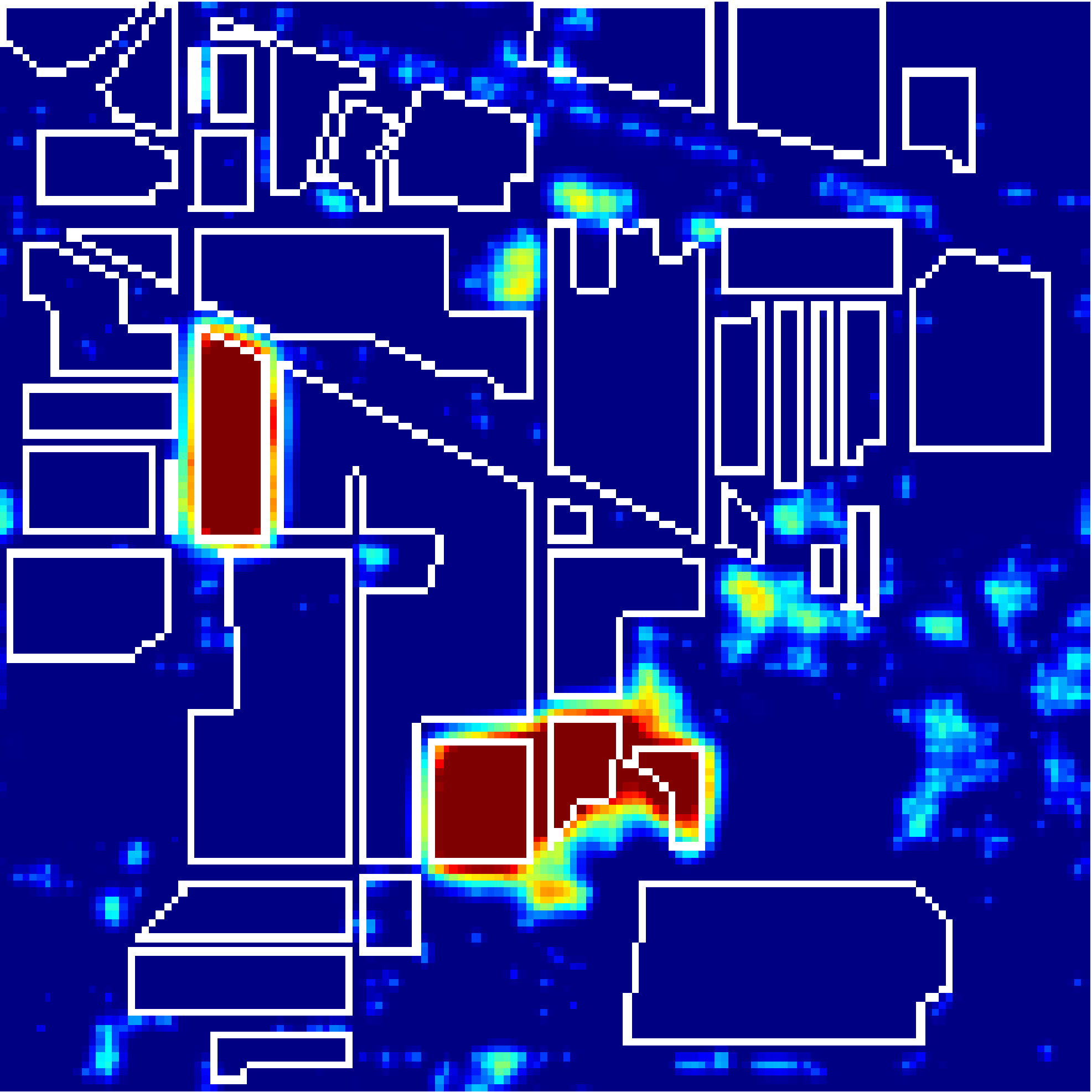} &
\includegraphics[height=.35\columnwidth,width=.37\columnwidth]{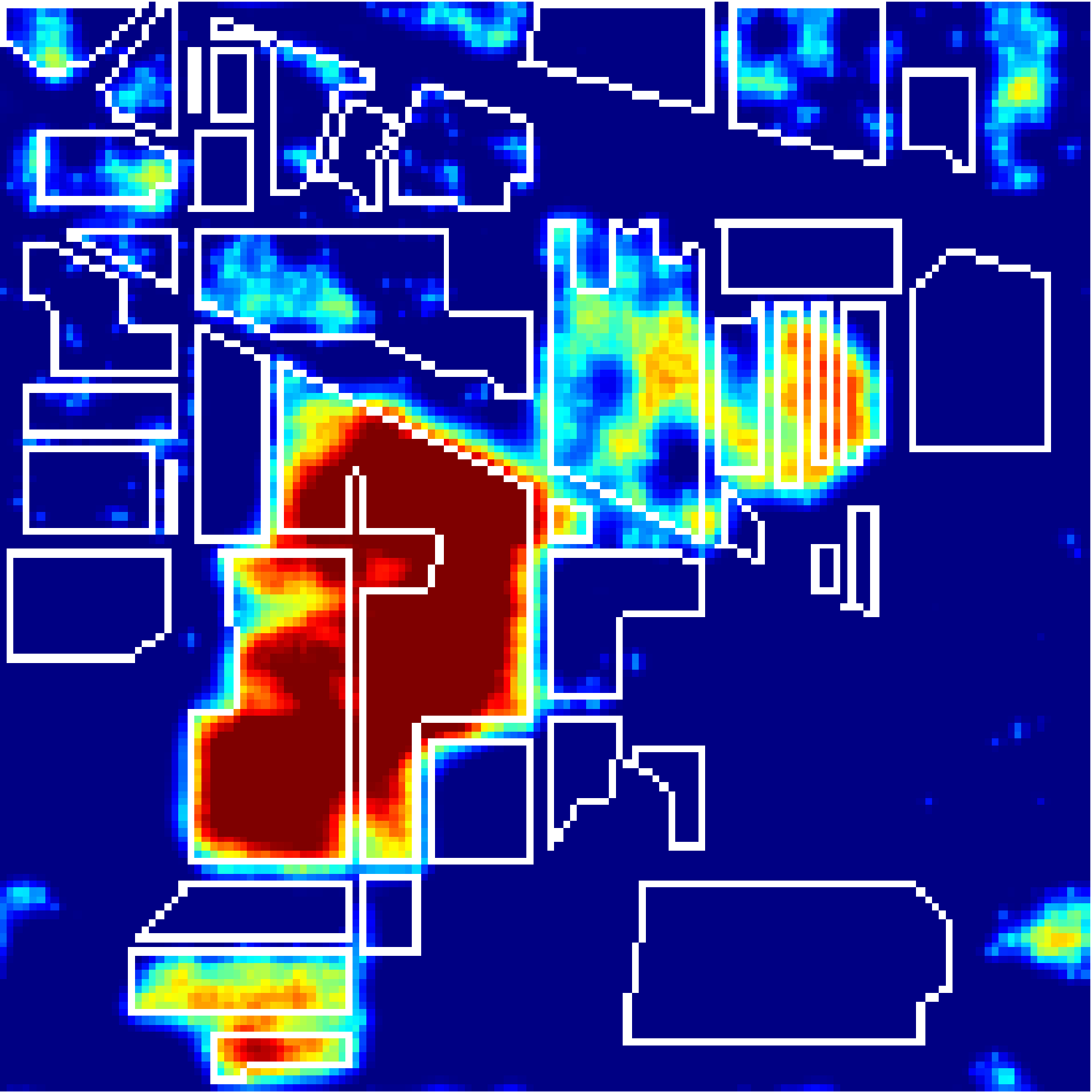} \\
($a$)& ($b$)  & ($c$) & ($d$) & ($e$)\\
\includegraphics[height=.35\columnwidth,width=.37\columnwidth]{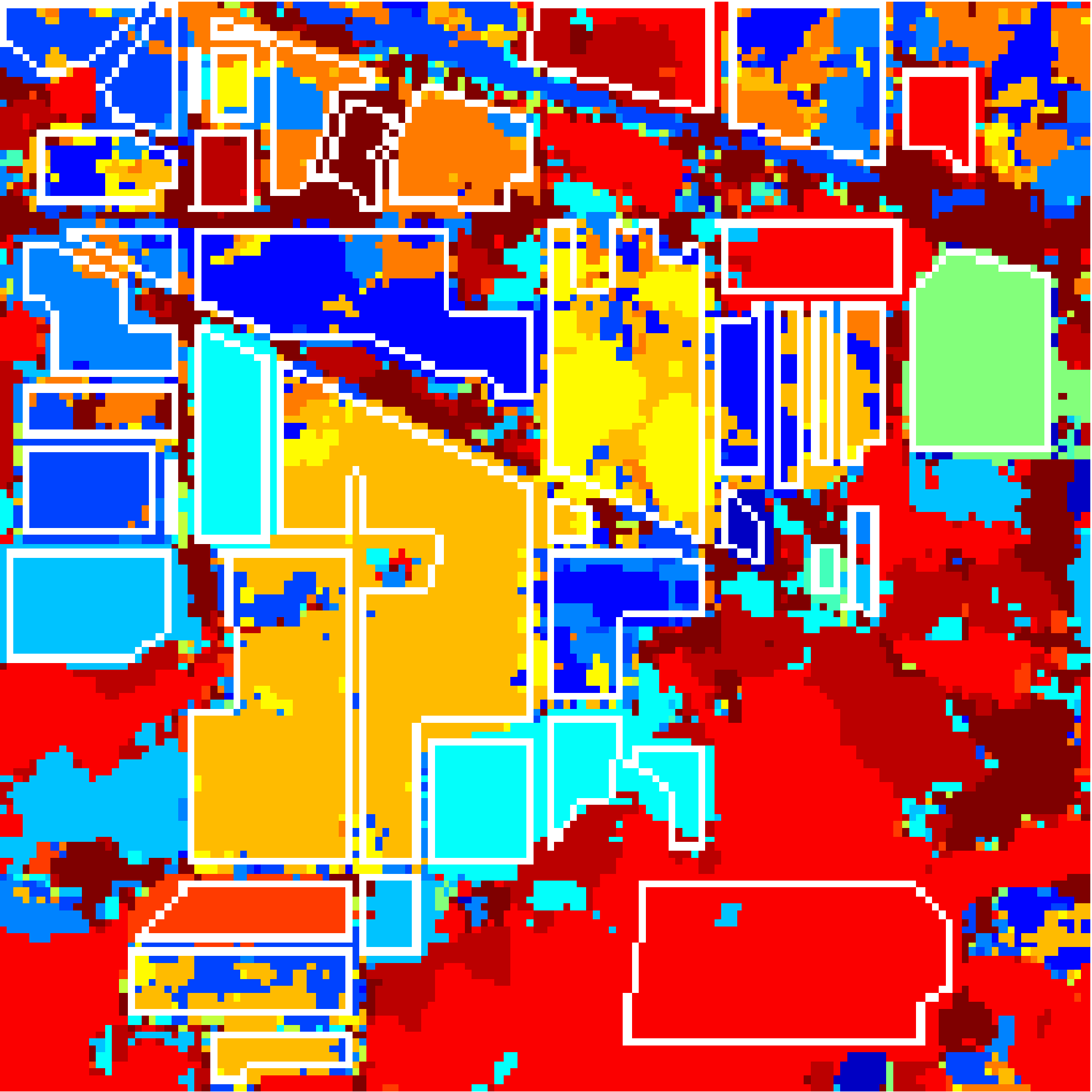} &
\includegraphics[height=.35\columnwidth,width=.37\columnwidth]{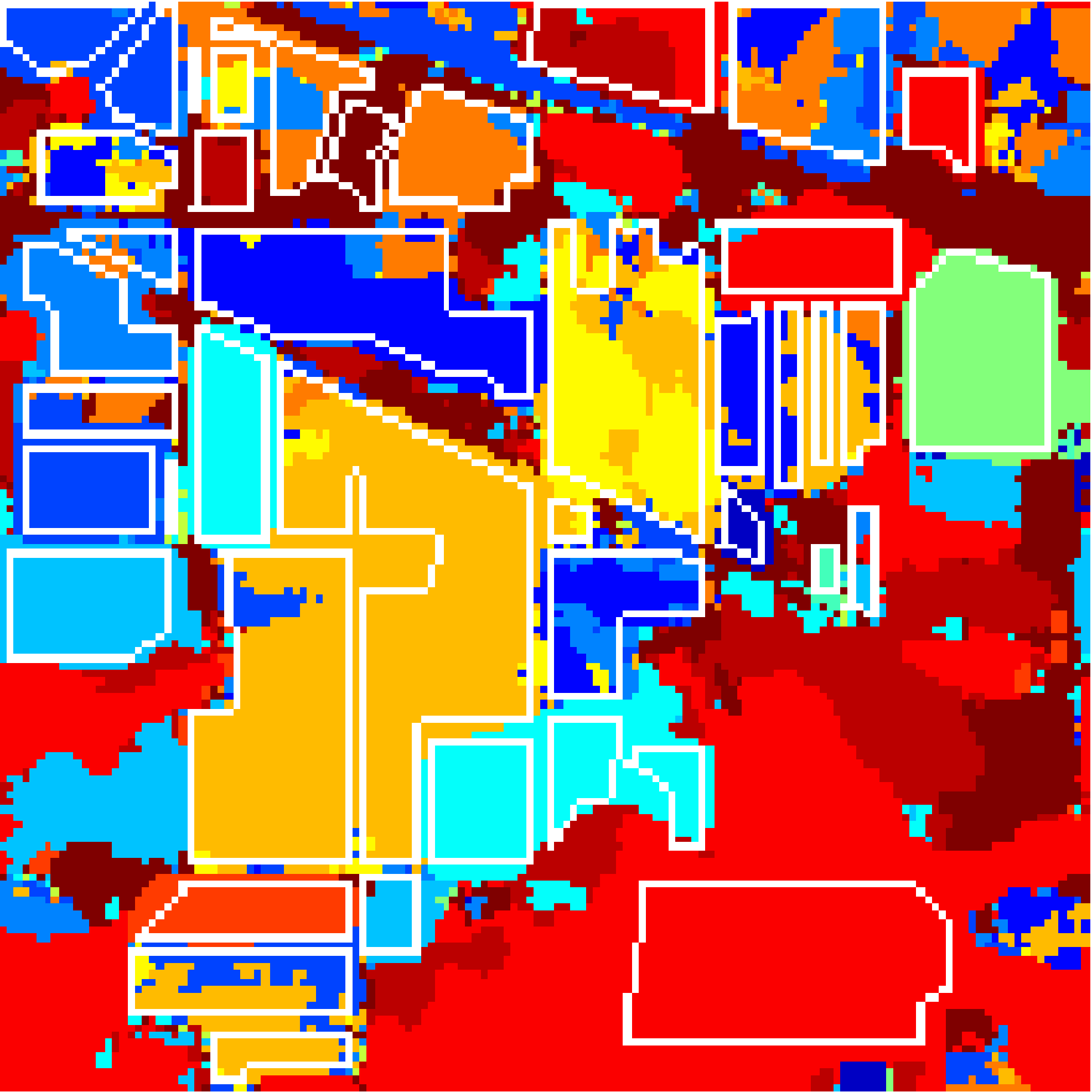} &
\includegraphics[height=.35\columnwidth,width=.37\columnwidth]{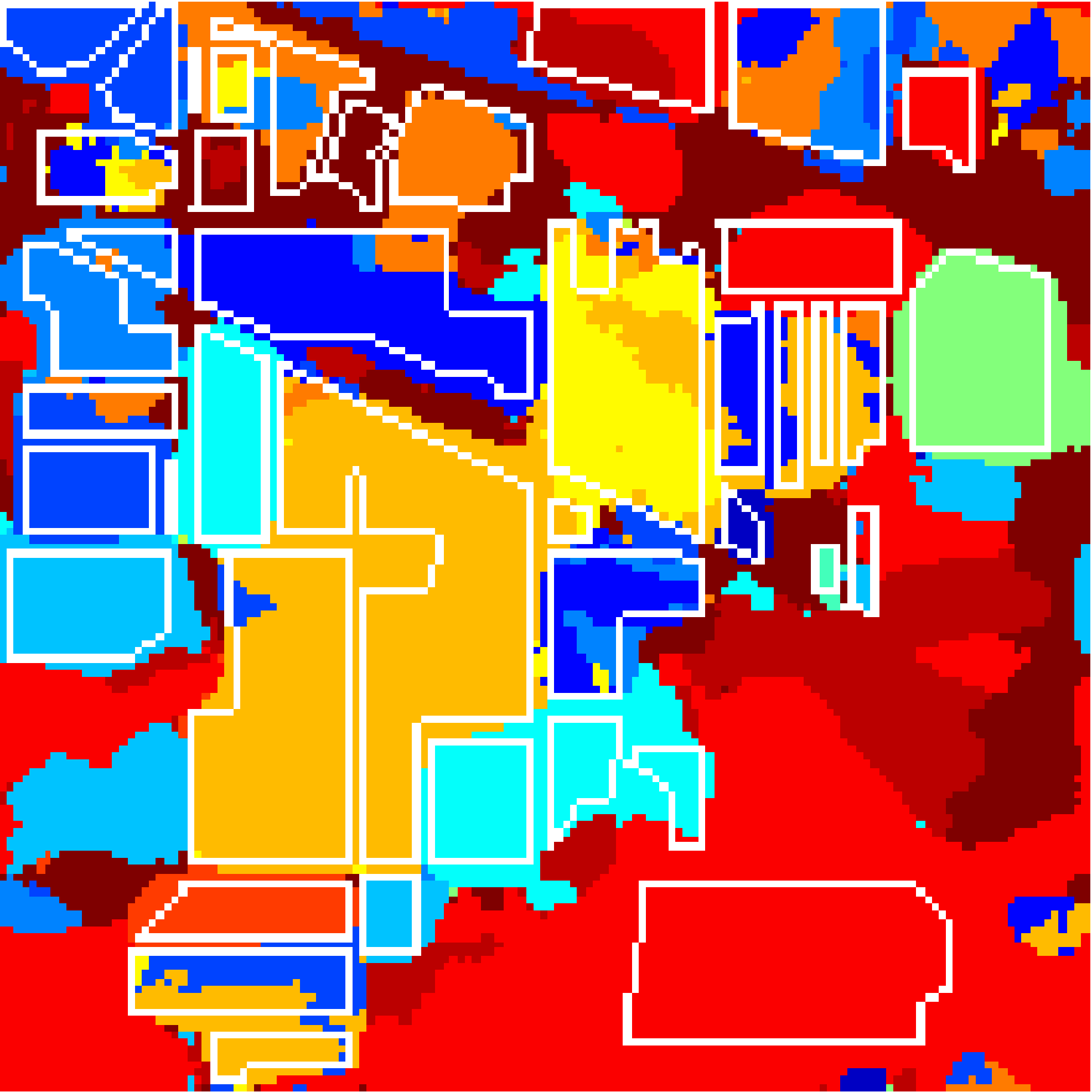} &
\includegraphics[height=.35\columnwidth,width=.37\columnwidth]{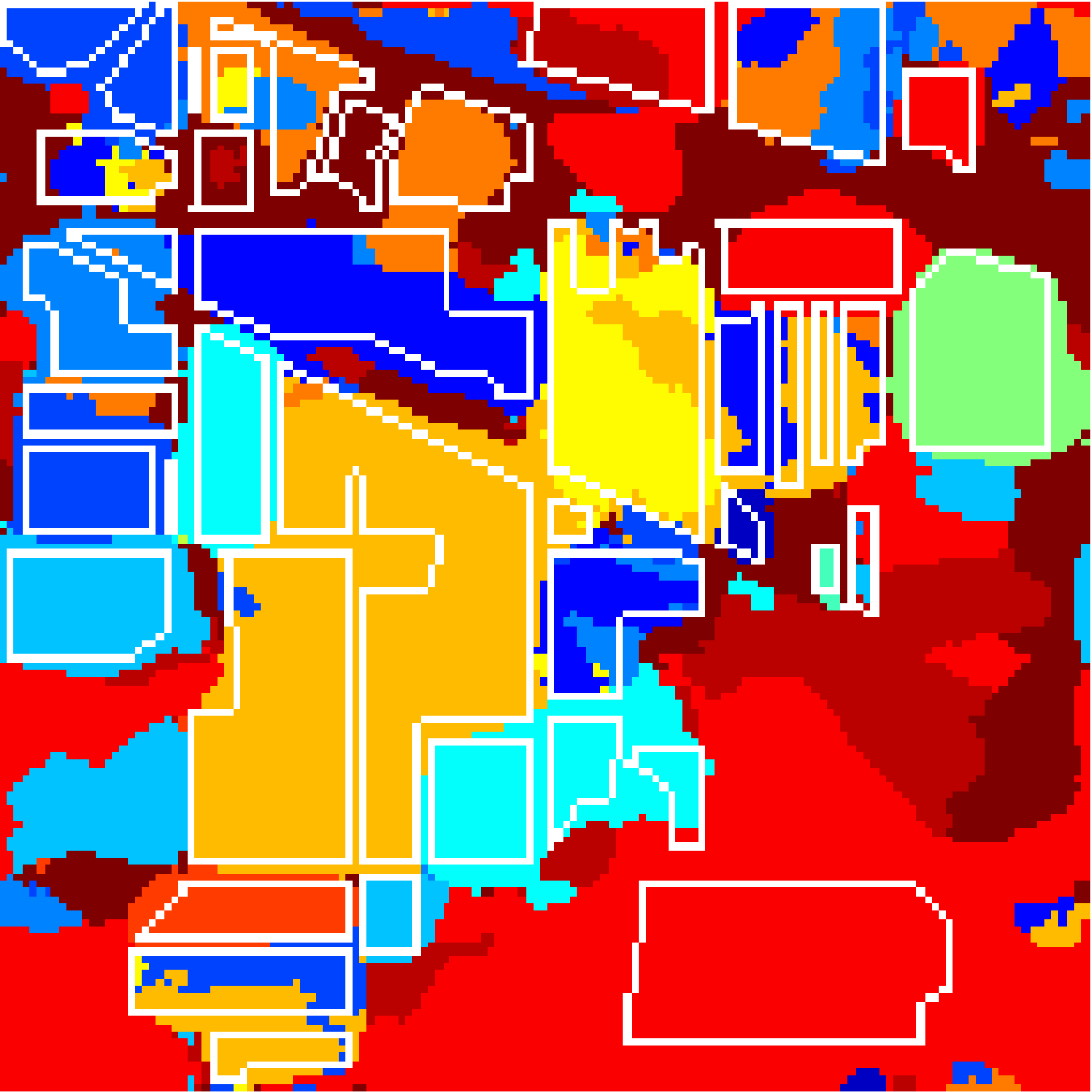} &
\includegraphics[height=.35\columnwidth,width=.37\columnwidth]{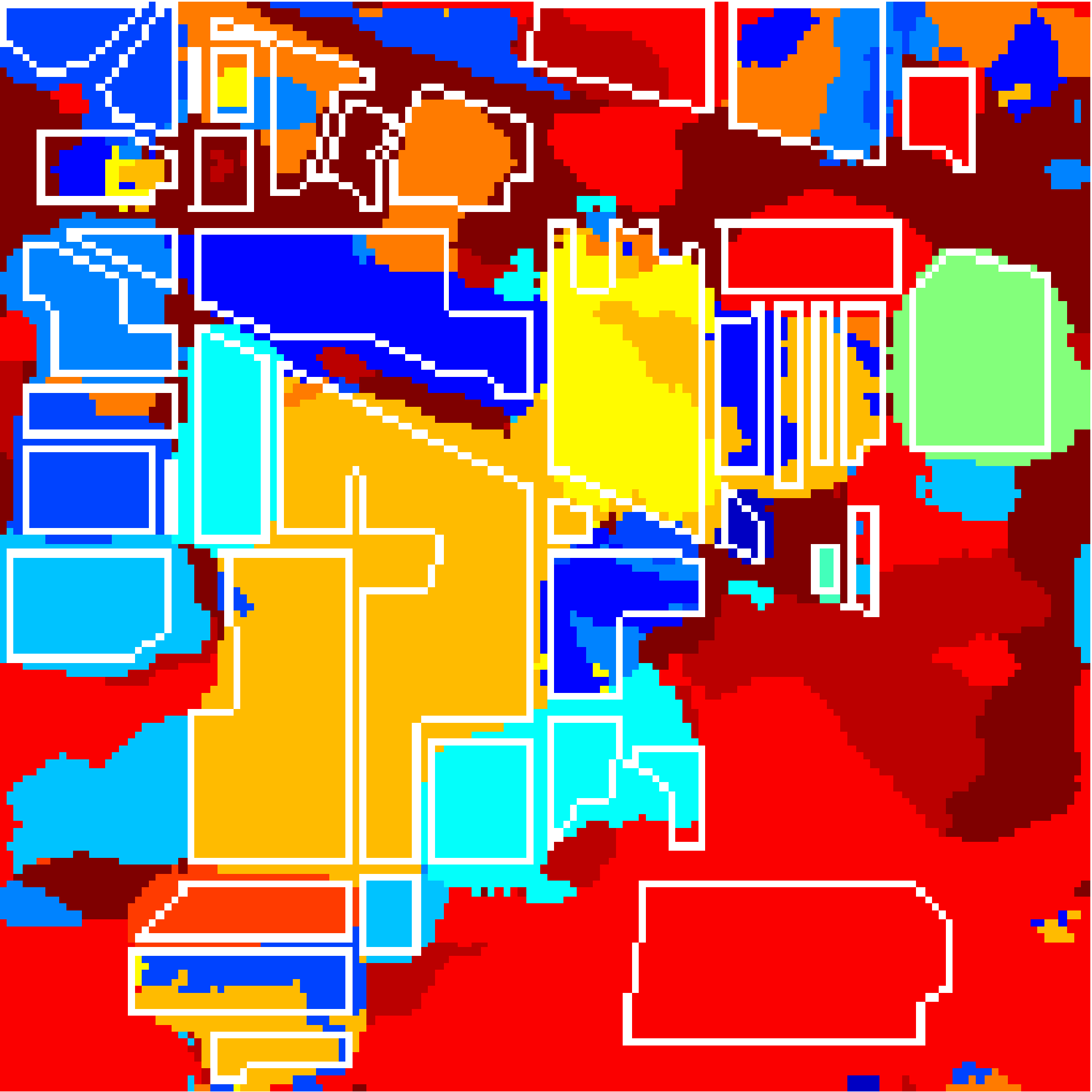} \\
 ($f$) & ($g$) & ($h$) & ($i$) & ($j$)\\
\end{tabular}
\caption{\small \label{fig:results}
AVIRIS Indian Pines scene. SegSALSA-STR segmentation results for multiple patch sizes, starting from the same MLR segmentation.
Top row --- ($a$) ground truth, ($b$) MLR segmentation -- $56.86\%$ accuracy, ($c$) segmentation for $5 \times 5$ patches -- $78.60\%$ accuracy, ($d$) hidden field for grass-trees class (light blue) for segmentation ($c$), ($e$) hidden field for soybean-mintill class (dark yellow) for segmentation ($c$).
Bottom row --- ($f$) segmentation for $1 \times 1$ patches (corresponding to SegSALSA with VTV priors) -- $74.33 \%$ accuracy, ($g$) segmentation for $3 \times 3$ patches -- $77.53 \%$ accuracy, ($h$) segmentation for $7 \times 7$ patches -- $78.64 \%$ accuracy, (i) segmentation for $9 \times 9$ patches -- $78.33 \%$, ($j$) segmentation for $11 \times 11$ patches -- $ 78.01 \%$ accuracy.
}
\end{center}
\end{figure*}

\paragraph*{Complexity, parallelization and stopping criterion}
The time complexity of solving the problem \eqref{eq:u_update} is dominated by the computation of $n$ single value decompositions of matrices $( L K) \times 2$, which amounts to a time complexity of $O(n (LK)^3)$~\cite{GolubV:89}. 
The complexity of SegSALSA-STR is 
\begin{equation*}
O(L K n \log n + n (LK)^3),
\end{equation*}
being highly parallelizable: $L K$ decoupled fast Fourier transforms followed by $n$ decoupled SVDs.
As a stopping criterion, we impose that both the primal and dual residuals are smaller than a given threshold.
It was observed that a fixed number of iterations of the order of $100$ provides excellent results.

\section{Results}\label{sec:results}
We validate our algorithm in the segmentation of a subscene of the AVIRIS Indian Pine scene (Fig. \ref{fig:results}).
The subscene consists of a $145\times 145$ pixel subsection with $200$ spectral bands and contains $16$ different classes. 
$15$ Samples per class are used as training set.
The classes are modeled by a multinomial logistic regression (MLR) and the MLR weights are learnt using the LORSAL~\cite{LiBDP:11}.
The parameter $\lambda$ corresponding to the relative weight \eqref{eq:problem} of the structure tensor prior is set to $2$, a compromise that allows good classification performance with a lower number of iterations of the algorithm.

We illustrate the joint effect of the size of the patch and of the bandwidth $\gamma$ of the Gaussian weighting for the construction of the patch-based Jacobian \eqref{eq:patch}, by comparing the segmentation results of a real hyperspectral image (Fig. \ref{fig:results}) for varying sizes of the patches.
The bandwith $\gamma$ is set as the distance in pixels from the border of the patch to the center (\emph{e.g.} $\gamma = 2$ for $5\times 5$ patches), with $\gamma =1 $ for $1\times 1$ patches.
As structure tensor regularization is a generalization of the VTV prior, in the case of $1 \times 1$ patches, SegSALSA-STR corresponds to SegSALSA with the VTV prior~\cite{BioucasDiasCK:14,CondessaBDK:14}.

The analysis of the segmentation results by SegSALSA-STR shows the capabilities of the structure tensor prior in obtaining smooth segmentation boundaries.
The hidden fields corresponding to two different classes (Figs. $d$, $e$) illustrate the preservation of detail of the segmentation (Fig. $c$).
The effect of increasing the patch size is evident (bottom row) as the segmentation performance for higher patch sizes is higher than the segmentation performance for lower patch sizes.

The use of structure tensor priors with a patch size larger than $1\times 1$  (Figs. $c$, $g$, $h$, $i$), allows an increase of segmentation performance when compared to $1\times 1$ patches (Fig. $f$).
This shows that extending the SegSALSA algorithm by using a generalization of the VTV prior based on structure tensor regularization allows for improved segmentation performance at the cost of a higher computational burden.

\section{Concluding Remarks}
\label{sec:conclusion}
In this paper we extended the SegSALSA algorithm to use a generalized total variation prior based on structure tensor regularization --- SegSALSA-STR.
This is a more general formulation of the SegSALSA algorithm which
extends the concept of VTV from the pixel level to the patch level.
The shift from a discrete to a continuous formulation paves the way
for a class of relevant approaches in remote sensing through the
inclusion of different priors.
The algorithm was validated in simulated and real hyperspectral data.

\paragraph*{Acknowledgements}
The authors would like to thank D. Landgrebe at Purdue University for providing the AVIRIS Indian Pines scene.

\bibliography{}
\bibliographystyle{IEEEbib}
\end{document}